\documentclass[10pt, a4paper]{article}
\usepackage{algorithm}
\usepackage{algpseudocode}
\usepackage{enumitem}
\usepackage{multirow}
\usepackage{amsmath}
\usepackage{subcaption}
\usepackage{booktabs}
\usepackage[autostyle=true]{csquotes}
\usepackage[font=small,labelfont=bf,justification=centering]{caption}
\usepackage{polyglossia}
\setdefaultlanguage{english}
\setotherlanguage{bengali}

\newfontfamily\bengalifont[
  Script=Bengali,
  Path= ./ ,
  Extension=.ttf
]{Kalpurush}

\usepackage[final]{lrec2026} 

\title{BiST: A Gold Standard Bangla-English Bilingual Corpus for Sentence Structure and Tense Classification with Inter-Annotator Agreement}

\name{
\begin{tabular}{c}
\textbf{Abdullah Al Shafi$^{1,†}$, Swapnil Kundu Argha$^{1}$, M. A. Moyeen$^{1}$, Abdul Muntakim$^{2}$,} \\
\textbf{Shoumik Barman Polok$^{1}$}
\end{tabular}
}

\address{
$^{1}$ Department of Computer Science and Engineering, Khulna University of Engineering \& Technology \\
$^{2}$ Department of Computer Science, Kennesaw State University, United States \\
\ abdullah@iict.kuet.ac.bd, swapnilkundu01@gmail.com, moyeen.kuet@gmail.com, \\ amuntaki@students.kennesaw.edu, polokbarman874@gmail.com\\
$^{†}$Corresponding author
}

\abstract{
High-quality bilingual resources remain a critical bottleneck for advancing multilingual NLP in low-resource settings, particularly for Bangla. To mitigate this gap, we introduce BiST, a rigorously curated Bangla-English corpus for sentence-level grammatical classification, annotated across two fundamental dimensions: syntactic structure (Simple, Complex, Compound, Complex-Compound) and tense (Present, Past, Future). The corpus is compiled from open-licensed encyclopedic sources and naturally composed conversational text, followed by systematic preprocessing and automated language identification, resulting in 30,534 sentences, including 17,465 English and 13,069 Bangla instances. Annotation quality is ensured through a multi-stage framework with three independent annotators and dimension-wise Fleiss’ Kappa ($\kappa$) agreement, yielding reliable and reproducible labels with $\kappa$ values of 0.82 and 0.88 for structural and temporal annotation, respectively. Statistical analyses demonstrate realistic structural and temporal distributions, while baseline evaluations show that dual-encoder architectures leveraging complementary language-specific representations consistently outperform strong multilingual encoders. Beyond benchmarking, BiST provides explicit linguistic supervision that supports grammatical modeling tasks, including controlled text generation, automated feedback generation, and cross-lingual representation learning. The corpus establishes a unified resource for bilingual grammatical modeling and facilitates linguistically grounded multilingual research.
 \\ \newline }

\begin{document}
\maketitleabstract
\pagestyle{empty}

\section{Introduction}
English and Bangla (Bengali) are two of the most widely spoken languages in the world, spoken by more than 1.5 billion people around the globe\footnote{\label{fn:shared}\url{https://en.wikipedia.org/wiki/English\_language}}\textsuperscript{,}\footnote{\url{https://en.wikipedia.org/wiki/Bengali\_language}}. They are thus of utmost importance in both international communication and computational linguistics. English is universally recognized and widely spoken both as a native and a second language\footnotemark[1]. On the other hand, Bangla, with around 242 million native speakers, is the sixth most spoken native language, and the official language of Bangladesh and one of the recognized languages in India\footnotemark[2]. Although widely used and recognized all over the world, there is a crucial shortage concerning bilingual annotations of the two languages for grammatical properties of sentences \cite{alam2025bnsentmix}.

Multilingual corpora are becoming progressively more useful for the NLP community, since many practical tasks require the involvement of more than one language \cite{seto2025training, tong2025novelcr}, but high-quality bilingual labeled corpora are still not plentiful, especially for low-resource language scenarios \cite{alam2025bnsentmix, hossain2025recognition}. Simultaneously, sentence-level linguistic information plays a crucial role in a wide range of NLP applications, for example, syntactic analysis, sentence categorization, tense identification, educational NLP, and cross-lingual transfer learning \cite{ghosh2025hate, hasan2025emotion, kabir2023banglabook}. Moreover, it is important to understand the diverse ways that the structural, and temporal properties of sentences are manifested across languages and how models might integrate high-quality linguistics along with high-quality accuracy \cite{hasan2025emotion, kabir2023banglabook}. Research on Bangla NLP also reports that, despite wide usage, the lack of large-scale and well-structured corpora remains a major challenge for building strong models and conducting deeper linguistic studies\cite{hossain2025recognition}. The existing literature on Bangla NLP has focused largely on monolingual resources \cite{kabir2023banglabook, shafi2025structured}. In addition, some prior works have explored multilingual resources \cite{majumdar2022bengali}, but there remains a scarcity of bilingual corpora that provide unified and explicitly annotated grammatical information across languages, particularly for dual-dimensional Bangla-English bilingual settings. This limits direct comparison of linguistic properties across two languages in one benchmark. To address this gap, we introduce BiST,a large-scale Bangla-English bilingual corpus comprising 30,534 sentences, annotated for two fundamental grammatical dimensions, namely sentence structure and tense. 



\begin{figure*}[ht]
\centering
\centerline{\includegraphics[scale=.34]{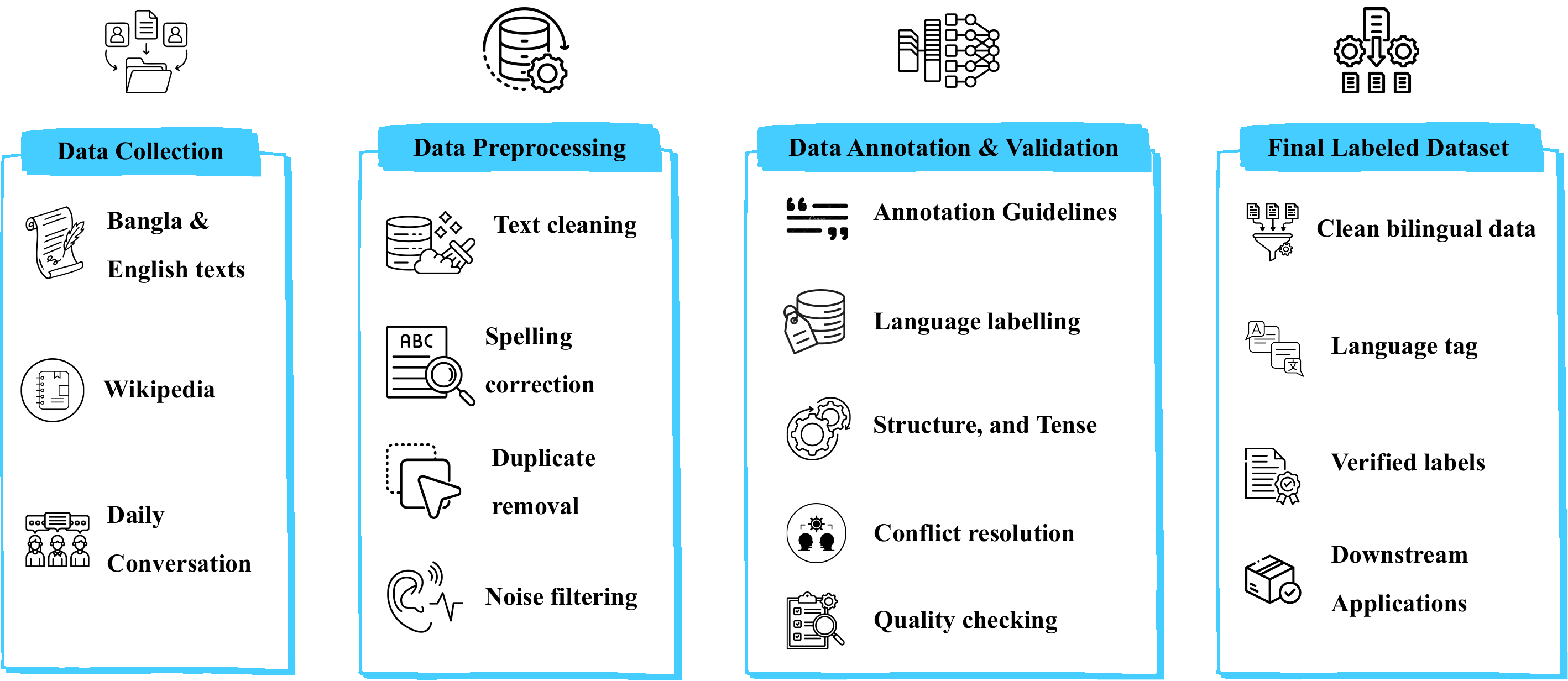}} 
\caption{Flowchart of data collection and annotation of our proposed \textquote*{BiST} Corpus.}
\label{fig:proposed}
\end{figure*}

\vspace{-1em}
\subsection{Contributions}
The main contributions of this work are as follows:

(i) To the best of our knowledge, BiST is the first Bangla-English bilingual corpus annotated jointly for sentence structure and tense with validated inter-annotator agreement.

(ii) A systematic multi-stage annotation framework is designed involving multiple annotators, majority voting, and consensus-based resolution, ensuring high-quality labels with strong inter-annotator agreement (IAA).

(iii) Explicit and standardized annotation guidelines are provided for both structural and temporal classification, enabling consistency and reproducibility in future research.

(iv) We establish baseline performance benchmarks using both multilingual encoders and dual-encoder architectures, demonstrating the effectiveness of language-specific representations.

(v) The corpus\footnote{\url{https://github.com/AbdullahRatulk/BiST}} is released as a publicly available resource to support reproducible research in bilingual and cross-lingual NLP.

The remainder of this paper is organized as follows. Section \ref{sec:corpus} describes the BiST corpus, including data collection, annotation, and its potential downstream applications. Section \ref{sec:result} presents the experimental analysis, including annotation reliability, statistical properties, and baseline evaluations. Finally, Section \ref{sec:conclusion} concludes the paper.

\section{A Bilingual Dual-Dimensional Sentence Classification Corpus}
\label{sec:corpus}
The workflow for the generation of the developed corpus is represented in Fig. \ref{fig:proposed}, illustrating the end-to-end process from data collection through preprocessing and language identification to multi-stage annotation with conflict resolution. The process results in a clean, reliably labeled bilingual corpus for downstream NLP tasks. 


\begin{algorithm}[t]
\caption{Bangla-English Word-Level Classifier}
\label{alg:bangla_english}
\begin{algorithmic}[1]
\Require Text string
\Ensure Language label: Bangla, or English

\State Initialize Bangla\_words $\gets 0$
\State Initialize English\_words $\gets 0$
\State Split text into words
\ForAll{words in text}
    \If{word contains Bangla characters}
        \State Bangla\_words $\gets$ Bangla\_words + 1
    \ElsIf{word contains English letters or digits}
        \State English\_words $\gets$ English\_words + 1
    \EndIf
\EndFor
\If{Bangla\_words $> 0$ AND English\_words $= 0$}
    \State \Return "Bangla"
\ElsIf{English\_words $> 0$ AND Bangla\_words $= 0$}
    \State \Return "English"
    
\EndIf
\end{algorithmic}
\end{algorithm}

\begin{algorithm}[t]
\caption{BiST Annotation with Dimension-wise Fleiss' Kappa}
\label{alg:bnengl_annotation}
\begin{algorithmic}[1]
\Require Sentences $S=\{s_1,\dots,s_n\}$
\Ensure Final labels 
$Y=\{(S_i,T_i)\}_{i=1}^{n}$
\State Define annotators $P=\{a_1,a_2,a_3\}$
\State Define label sets $\mathcal{S}, \mathcal{T}$
\ForAll{$s_i$ and $a_j$}
    \State Annotator assigns 
    $s_{ij}\in\mathcal{S}$,
    $t_{ij}\in\mathcal{T}$
\EndFor
\For{$d \in \{\mathcal{S},\mathcal{T}\}$}
    \State Compute item agreement $A_i^{(d)}$
    \State $\bar{A}^{(d)}=\frac{1}{n}\sum_{i=1}^{n} A_i^{(d)}$
    \State Compute expected agreement $A_e^{(d)}$
    \State Fleiss' Kappa, $\kappa^{(d)}=\frac{\bar{A}^{(d)}-A_e^{(d)}}{1-A_e^{(d)}}$
\EndFor

\ForAll{$s_i$}
    \State $S_i \gets \underset{s \in \mathcal{S}}{\operatorname{arg\,max}} \sum_{j=1}^{3} \mathbf{1}(s_{ij} = s)$
    \State $T_i \gets \underset{t \in \mathcal{T}}{\operatorname{arg\,max}} \sum_{j=1}^{3} \mathbf{1}(t_{ij} = t)$
    \If{no majority in any dimension}
        \State Conduct consensus discussion
    \EndIf
\EndFor

\Return Final labeled corpus $Y$
\end{algorithmic}
\end{algorithm}

\begin{table*}[h]
\centering
\begin{tabular}{|l|p{6cm}|p{8cm}|}
\hline
\textbf{Attribute} & \textbf{Description and possible values} & \textbf{Example} \\
\hline
Sentence & The full Bangla or English sentence collected from raw sources. Each sentence occupies one row. 
& \textbengali{১৯৫২ সালের ২১শে ফেব্রুয়ারিতে এই আন্দোলন চূড়ান্ত রূপ ধারণ করলেও বস্তুত এর বীজ রোপিত হয়েছিল বহু আগে; অন্যদিকে এর প্রতিক্রিয়া এবং ফলাফল ছিল সুদূরপ্রসারী। (Although this movement took its final form on February 21, 1952, its seeds were actually planted much earlier; on the other hand, its reactions and consequences were far-reaching.)}\\
\hline
Language & Indicates the language of the sentence (English/Bangla). & Bangla\\
\hline
Structure & The grammatical sentence structure, determined by clause composition. Possible values include: Simple, Complex, Compound, and Complex-compound. 
& Complex-compound \\
\hline
Tense & The primary tense expressed in the sentence. Includes major tense categories such as Present, Past, and Future. 
& Past \\
\hline
\end{tabular}
\caption{corpus description with attributes and possible values.}
\label{description}
\end{table*}

\subsection{Data Collection}
The very first step is to collect the documents composed in Bangla and/or English from a variety of sources. Sentences were collected from two sources: (i) Wikipedia\footnote{\url{https://en.wikipedia.org/wiki/}}\textsuperscript{,}\footnote{\url{https://bn.wikipedia.org/wiki/}} and (ii) naturally occurring everyday conversational texts. Sentences were collected from Wikipedia in accordance with its open licensing policy (Creative Commons Attribution-ShareAlike license), which explicitly permits reuse and redistribution under the specified terms. Wikipedia data were collected using automated web crawling techniques implemented through BeautifulSoup\footnote{\url{https://beautiful-soup-4.readthedocs.io/en/latest/}}, strictly adhering to Wikipedia’s Terms of Service and robots.txt guidelines. The conversational portion of the corpus consists of manually composed and curated sentences designed to reflect realistic informal communication patterns, including dialogue-style expressions, question-answer formats, and everyday interactions like requests, opinions, and narratives. These conversational instances do not originate from copyrighted or restricted sources, and therefore do not impose redistribution constraints.

\begin{figure*}[htbp]
\centering
     \begin{subfigure}[b]{0.44\textwidth}
         \centering
         \includegraphics[width=\textwidth]{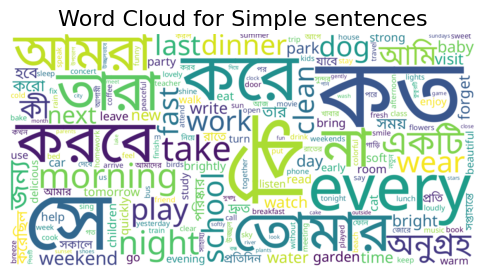}
         \caption{Simple}
         \label{wc_simple}
     \end{subfigure}
     \begin{subfigure}[b]{0.44\textwidth}
         \centering
         \includegraphics[width=\textwidth]{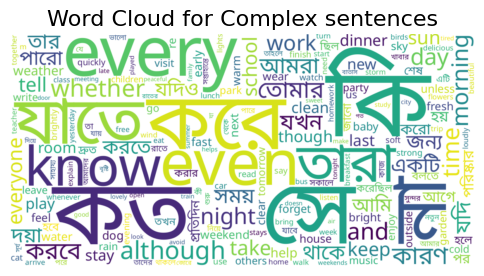}
         \caption{Complex}
         \label{wc_complex}
     \end{subfigure}
      \begin{subfigure}[b]{0.44\textwidth}
         \centering
         \includegraphics[width=\textwidth]{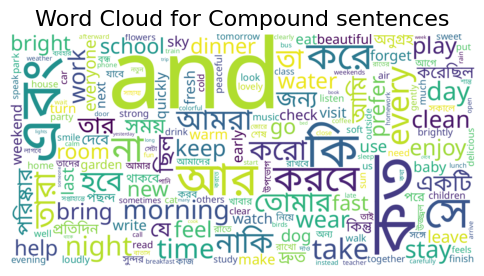}
         \caption{Compound}
         \label{wc_compound}
     \end{subfigure}
     \begin{subfigure}[b]{0.49\textwidth}
         \centering
         \includegraphics[width=\textwidth]{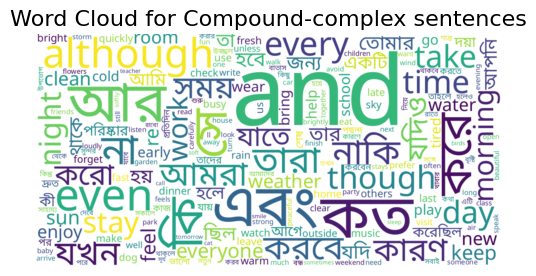}
         \caption{Complex-compound}
         \label{wc_cc}
     \end{subfigure}
        \caption{Word clouds representing the distribution of words across different \textquote*{structure} types in the corpus.}
\label{wcs}
\end{figure*}

\vspace{-1em}
\subsection{Annotation Guidelines}
To ensure consistency and reproducibility, explicit annotation guidelines were established prior to the labeling process. Annotators were instructed to assign labels based on clause composition and primary temporal reference, following standardized linguistic definitions.

\textbf{Sentence Structure Classes:} Sentence structure is determined based on the number and type of clauses:

(1) \textbf{Simple:} A sentence containing a single independent clause. For example, \textquote*{\textbengali{বর্তমানে ২,১১২ জন সক্রিয় স্বেচ্ছাসেবক বাংলা উইকিপিডিয়ায় কাজ করছেন। (Currently, 2,112 active volunteers are working on Bangla Wikipedia.)}}, \textquote*{The boy with glasses was looking at the moon.}, etc.

(2) \textbf{Complex:} A sentence with one independent clause and at least one subordinate clause. Some examples include \textquote*{\textbengali{সে প্রতিদিন সন্ধ্যায় খবর দেখে যাতে সে বর্তমান ঘটনার সম্পর্কে আপডেট থাকে। (He watches the news every evening so that he can stay updated on current events.)}}, \textquote*{Do they enjoy camping even if the weather is bad?}, etc.

(3) \textbf{Compound:} A sentence consisting of two or more independent clauses joined by coordinating conjunctions. For instance, \textquote*{\textbengali{তুমি কি গিটার বাজাও, নাকি অন্য কোনো বাদ্যযন্ত্র বাজাও? (Do you play the guitar or any other musical instrument?)}}, \textquote*{He wears a suit, but he prefers casual clothes.}, etc.

(4) \textbf{Complex-compound:} A sentence containing multiple independent clauses along with at least one subordinate clause. For example, \textquote*{\textbengali{তোমার কাপড় গুছিয়ে ভাঁজ করে রাখো, আর ঠিকঠাক জায়গায় রেখে দাও যাতে ঘর পরিপাটি দেখায়। (Fold your clothes and put them in their proper place so that the house looks tidy.)}}, \textquote*{She reads books as she enjoys learning, and shares her notes with others.}, etc.


\textbf{Sentence Tense Classes:} Tense annotation focuses on the primary temporal reference of the main clause:

(1) \textbf{Present:} Actions occurring currently or habitually. For example, \textquote*{\textbengali{আমি কাজ করার সময় গান শুনি কারণ এটা মনোযোগ ধরে রাখতে সাহায্য করে, এবং এমন প্লেলিস্ট বেছে নিই যা আমাকে উদ্দীপ্ত রাখে। (I listen to music while I work because it helps me stay focused, and I choose playlists that keep me motivated.)}}, \textquote*{Help your grandparents when they need assistance.}, etc.

(2) \textbf{Past:} Actions completed in the past. For instance, \textquote*{\textbengali{এর প্রতিক্রিয়া এবং ফলাফল ছিল সুদূরপ্রসারী। (Its reaction and consequences were far-reaching.)}}, \textquote*{We watched the sunrise together while drinking hot coffee.}, etc.

(3) \textbf{Future:} Actions expected to occur in the future. Some examples include \textquote*{\textbengali{পার্টি কোথায় হবে? (Where will the party held?)}}, \textquote*{I will visit the museum tomorrow unless it is closed, and I will take many pictures.}, etc.

In cases of ambiguity, annotators prioritized the main clause for tense assignment and relied on clause hierarchy for structural classification.


\begin{figure*}[htbp]
\centering
     \begin{subfigure}[b]{0.44\textwidth}
         \centering
         \includegraphics[width=\textwidth]{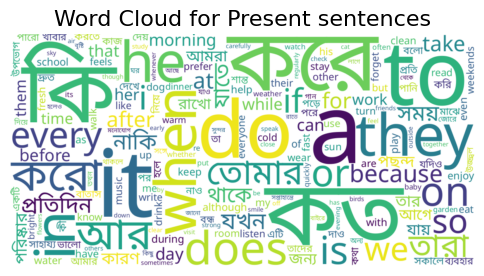}
         \caption{Present}
         \label{wc_present}
     \end{subfigure}
     \begin{subfigure}[b]{0.44\textwidth}
         \centering
         \includegraphics[width=\textwidth]{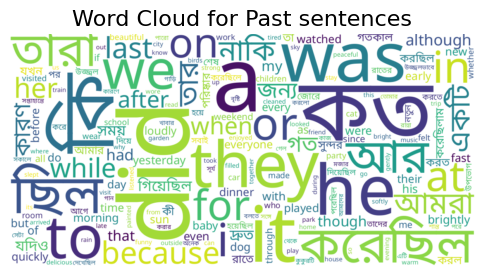}
         \caption{Past}
         \label{wc_past}
     \end{subfigure}
      \begin{subfigure}[b]{0.44\textwidth}
         \centering
         \includegraphics[width=\textwidth]{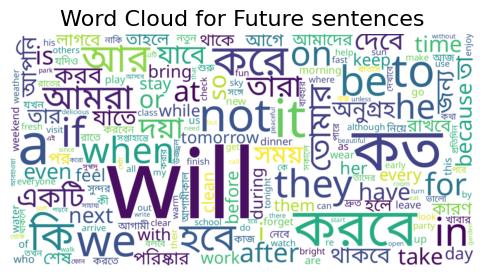}
         \caption{Future}
         \label{wc_future}
     \end{subfigure}
        \caption{Word clouds representing the distribution of words across different \textquote*{tense} categories in the corpus.}
\label{wct}
\end{figure*}

\subsection{Corpus Annotation}
After collecting data and defining annotation guidelines, the compilation then goes through a series of very extensive preprocessing steps such as cleaning, spelling correction, removing duplicates, and filtering noise to assure the quality of the data. After preprocessing, the language tag was generated automatically through python script leveragin Algotithm \ref{alg:bangla_english}. After that, the corpus was annotated in a multi-stage verification setting to ensure consistency and reliability as described in Algorithm \ref{alg:bnengl_annotation}. The proposed BiST annotation framework systematically assigns final labels for each sentence along two linguistic dimensions: structure ($S$), and tense ($T$). Initially, a set of sentences $\mathcal{S} = \{ s_1, \dots, s_n \}$ is prepared, and three annotators $\mathcal{P} = \{ a_1, a_2, a_3 \}$ independently label each sentence according to the pre-defined label sets $S$, and $T$. For each sentence $s_i$ and annotator $a_j$, the assigned labels are denoted $s_{ij} \in S$, and $t_{ij} \in T$.

To ensure annotation reliability, dimension-wise agreement is computed using Fleiss’ Kappa. For each dimension $d \in \{S, T\}$, the item-wise agreement $A_i(d)$ is first calculated and then averaged across all sentences to obtain $\bar{A}(d)$. The expected agreement $A_e(d)$ is determined under the assumption of random labeling, and the dimension-specific Fleiss’ Kappa is computed. After measuring IAA, majority voting is applied to determine the final label for each sentence in all three dimensions. In cases where no clear majority exists in any dimension, a consensus discussion among annotators is conducted to finalize the label. The final output of the algorithm is a fully labeled corpus $\mathcal{Y} = \{ (S_i, T_i) \}_{i=1}^{n}$ which can be used for subsequent computational analyses, such as supervised classification or linguistic studies.


\begin{table*}[t]
\centering
\small
\begin{minipage}{0.48\linewidth}
\centering
\begin{tabular}{lcc}
\hline
\textbf{Dimension} & \textbf{Fleiss' Kappa ($\kappa$)} & \textbf{Agreement Level} \\
\hline
Structure & 0.82 & Almost Perfect \\
Tense & 0.88 & Almost Perfect \\
\hline
\end{tabular}
\caption{Inter-annotator agreement measured using Fleiss' Kappa across linguistic dimensions.}
\label{tab:fliess_kappa}
\end{minipage}
\hfill
\begin{minipage}{0.48\linewidth}
\centering
\begin{tabular}{lccc}
\toprule
\textbf{Statistic} & \textbf{Total} & \textbf{English} & \textbf{Bangla} \\
\midrule
Count & 30,534 & 17,465 & 13,069 \\
$\mu$ (Mean) & 10.18 & 10.8 & 9.36 \\
$\sigma$ (Std) & 4.48 & 4.66 & 4.07 \\
\bottomrule
\end{tabular}
\caption{Sentence length statistics of the corpus across English and Bangla sentences.}
\label{tab:length_stats}
\end{minipage}
\end{table*}

\subsection{Corpus Description and Visualization}
This subsection provides a clear description and visualization of the developed corpus, allowing researchers to effectively use this resource for bilingual NLP tasks and sentence-level linguistic analysis. Table \ref{description} outlines the attributes included in the corpus, describing each feature along with its possible values and illustrative examples. It outlines the linguistic, grammatical, and structural features of every sentence, thereby providing a clear picture of the corpus’s annotation scheme.

In addition, Each subfigure of Fig. \ref{wcs} depicts the core words corresponding to a distinct sentence structure. The differing levels of lexical prominence within the context of different categories help to qualitatively identify the differing sentence structures, such as the usage of conjunctions and clause markers in compound and complex sentences, respectively.

Furthermore, Fig. \ref{wct} identifies the tense-specific lexical features like temporal markers and auxiliary verbs, which occur in a more prominent manner in the respective categories. Variations in the occurrence of words in different tenses elucidate the efficacy of the corpus in encapsulating the temporal nuances in English as well as the Bengali sentences.


\subsection{Corpus Utility and Downstream Applications}
In addition to serving as a benchmark for sentence-level grammatical classification, the BiST corpus provides explicit linguistic supervision that supports a range of downstream NLP applications. Sentence structure and tense annotations are particularly useful for controlled text generation \cite{liang2024controllable}, where models are required to produce outputs with specific grammatical constraints. They also benefit educational NLP systems, including grammar correction \cite{fang2025llmcl}, automated feedback generation \cite{lindsay2025responsible}, and language learning applications. Furthermore, the availability of aligned annotations across Bangla and English facilitates cross-lingual transfer learning \cite{ma2025cross}, enabling models to better capture syntactic and temporal correspondences between languages.

Beyond these applications, the annotated structural and temporal features can enhance sentence-level representations for tasks such as sentiment analysis and stance detection \cite{shafi2026kuet}, where discourse structure and temporal cues often influence interpretation. These annotations are also valuable for information extraction \cite{li2025matching} and question answering \cite{zannat2025bridging}, where temporal understanding is essential for reasoning over events, as well as for text simplification and rewriting systems \cite{agrawal2024text}, which rely on structural information to transform complex sentences into simpler forms.

\section{Experimental Analysis and Result}
\label{sec:result}
This section provides a comprehensive evaluation of the proposed BiST corpus from both linguistic and computational perspectives. We begin by assessing annotation reliability through inter-annotator agreement. We then analyze the statistical properties and distributional characteristics of the corpus across structural and temporal dimensions. Finally, we establish baseline results using multilingual and dual-encoder architectures to demonstrate the utility of BiST for bilingual sentence classification tasks.
\vspace{-1em}
\subsection{Annotation Reliability}
Table \ref{tab:fliess_kappa} reports the dimension-wise inter-annotator agreement measured using Fleiss’ Kappa ($\kappa$) across the two annotated linguistic dimensions. The results indicate high reliability of the annotation framework. For sentence structure classification, a $\kappa$ value of 0.82 demonstrates strong agreement among annotators, reflecting consistent identification of clause composition patterns across Bangla and English sentences. Some disagreements occurred during annotation, particularly in distinguishing between complex, compound, and complex-compound sentences, where clause boundaries or relationships were not always explicitly marked. For tense classification, the agreement is even higher, with a $\kappa$ value of 0.88, indicating near-perfect consistency in identifying temporal categories. These scores confirm that the annotation guidelines were well-defined and systematically applied, and that the resulting labels are robust and suitable for benchmarking and supervised modeling. Overall, the high agreement values validate the quality and reproducibility of the BiST annotation process.



\begin{figure*}[htbp]
\centering
     \begin{subfigure}[b]{0.48\textwidth}
         \centering
         \includegraphics[width=\textwidth]{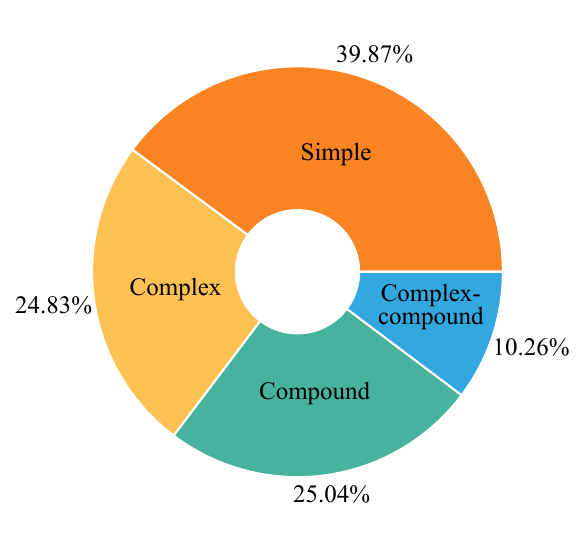}
         \caption{}
         \label{sd}
     \end{subfigure}
     \begin{subfigure}[b]{0.46\textwidth}
         \centering
         \includegraphics[width=\textwidth]{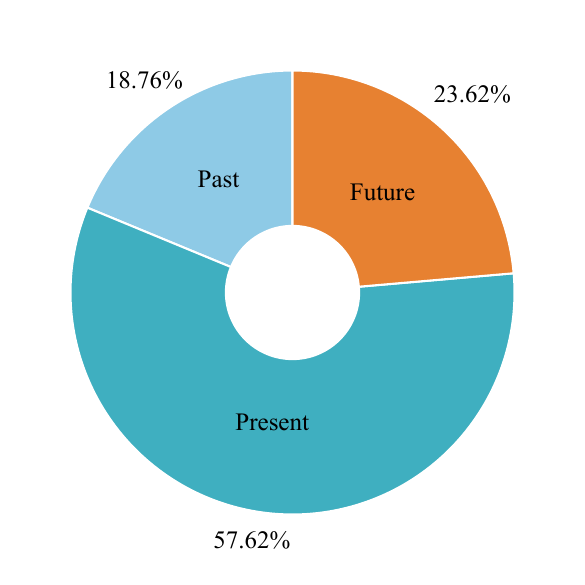}
         \caption{}
         \label{td}
     \end{subfigure}
        \caption{Distribution of (a) structural types and (b) temporal categories in the developed corpus.}
\label{std}
\end{figure*}

\begin{figure*}[htbp]
\centering
     \begin{subfigure}[b]{0.45\textwidth}
         \centering
         \includegraphics[width=\textwidth]{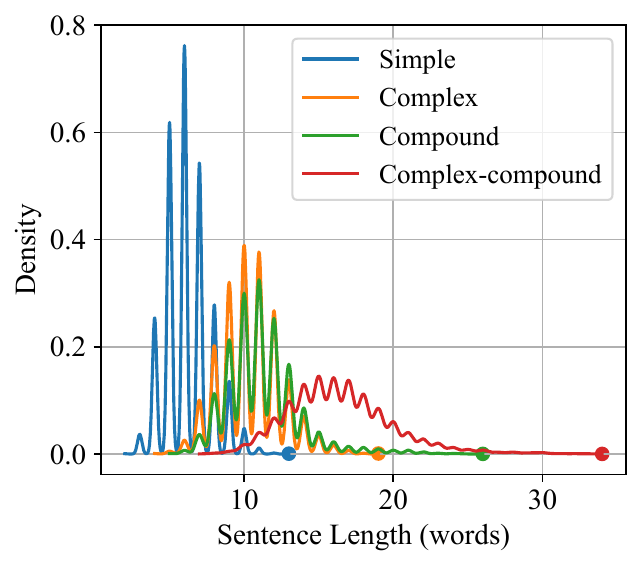}
         \caption{}
         \label{fig:skde}
     \end{subfigure}
     \begin{subfigure}[b]{0.44\textwidth}
         \centering
         \includegraphics[width=\textwidth]{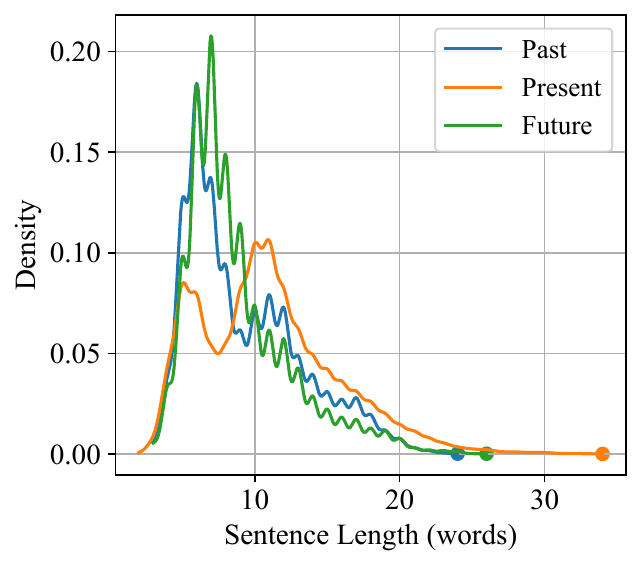}
         \caption{}
         \label{fig:tkde}
     \end{subfigure}
        \caption{Kernel density estimation (KDE) for (a) structural classes and (b) temporal categories in the developed corpus.}
\label{kde}
\end{figure*}

\begin{table*}[t]
\centering
\small
\resizebox{\textwidth}{!}{
\begin{tabular}{llccc ccc}
\hline
\multirow{2}{*}{Approach} & \multirow{2}{*}{Model} 
& \multicolumn{3}{c}{\textbf{Tense}} 
& \multicolumn{3}{c}{\textbf{Structure}} 
\\
\cmidrule(lr){3-5} \cmidrule(lr){6-8}
& & Acc & F1 & AUC 
& Acc & F1 & AUC \\
\hline
\multirow{2}{*}{Multilingual Encoders}
& mBERT 
& 0.85 & 0.83 & 0.90 
& 0.83 & 0.80 & 0.88 \\
& \textbf{XLM-R} 
& 0.87 & \textbf{0.85} & 0.91 
& 0.84 & \textbf{0.81} & 0.89 \\
\hline
\multirow{4}{*}{Dual Encoders}
& BERT + BanglaBERTBase 
& 0.87 & 0.85 & 0.91 
& 0.85 & 0.82 & 0.89 \\
& \textbf{BERT + BanglishBERT }
& \textbf{0.89} & \textbf{0.87} & \textbf{0.92} 
& \textbf{0.87} & \textbf{0.84} & \textbf{0.90} \\
& DistilBERT + BanglaBERTBase 
& 0.84 & 0.82 & 0.89 
& 0.84 & 0.81 & 0.88 \\
& DistilBERT + BanglishBERT 
& 0.86 & 0.84 & 0.90 
& 0.86 & 0.83 & 0.89 \\
\hline
\end{tabular}
}
\caption{Baseline evaluation with multilingual and dual-encoder models on structural and temporal classification tasks. Dual-encoder architectures consistently outperform multilingual baselines, indicating the advantage of combining complementary linguistic representations. Here, Acc = Accuracy, F1 = F1-Score, and AUC = Area Under the Curve.}
\label{tab:bilingual_results}
\end{table*}

\subsection{Statistical Analysis of the Corpus}
Table \ref{tab:length_stats} presents the sentence length statistics of the corpus, disaggregated by language. The corpus comprises a total of 30,534 sentences, including 17,465 English and 13,069 Bangla sentences, indicating a moderately higher representation of English data. In terms of sentence length, the overall average is $\mu = 10.18$ tokens per sentence. English sentences exhibit a higher average length ($\mu = 10.8$) compared to Bangla sentences ($\mu = 9.36$), suggesting that English instances in the corpus tend to be slightly more verbose. The standard deviation values further reveal variability patterns. English sentences show the highest dispersion ($\sigma = 4.66$), while Bangla sentences demonstrate comparatively lower variability ($\sigma = 4.07$). Thus, English sentence lengths are more widely distributed around the mean, whereas Bangla sentences are relatively more consistent in length. Overall, the statistics reflect a reasonably balanced bilingual corpus with moderate variation in sentence length across the two languages, while highlighting subtle structural differences between English and Bangla sentence construction.

Fig. \ref{std} presents the distribution of sentence types across structural categories and temporal categories within the developed corpus. In the structural distribution (Fig. \ref{sd}), simple sentences represent the largest proportion at 39.87\%, clearly exceeding the other categories. Compound (25.04\%) and complex (24.83\%) sentences account for nearly equal shares, while complex-compound sentences constitute the smallest proportion at 10.26\%. The dominance of simple sentences introduces a mild class imbalance, which may bias learning models toward simpler syntactic patterns unless balancing techniques such as weighting or augmentation are applied. From a linguistic perspective, the prevalence of simple sentences reflects natural language tendencies, where single-clause constructions are more common in everyday communication.

In the temporal distribution (Fig. \ref{td}), a more pronounced imbalance is evident. Present tense sentences account for 57.62\% of the corpus, more than double the proportion of past (18.76\%) and future (23.62\%) sentences. This skew mirrors real-world linguistic usage, where present tense dominates due to its applicability in general statements, habitual actions, and ongoing events. However, from a modeling standpoint, this imbalance may encourage classifiers to favor present-tense predictions, potentially reducing sensitivity to past and future constructions. The comparatively lower representation of future tense, in particular, may pose challenges for robust temporal classification.



Fig. \ref{fig:skde} presents the kernel density estimation (KDE) of sentence length distributions across four structural classes. The visualization reveals clear yet partially overlapping structural tendencies that offer both linguistic insight and modeling implications. A prominent observation is the strong concentration of simple sentences within shorter length ranges, typically peaking between 6-10 words. This aligns with their syntactic nature, as simple sentences generally consist of a single independent clause with minimal expansion. The narrow and sharply peaked density suggests low variance, indicating structural compactness and consistency. In contrast, complex sentences exhibit a moderate rightward shift in distribution, clustering around 9-13 words. This reflects the inclusion of subordinate clauses, which introduce additional syntactic material while maintaining a relatively controlled length. However, the overlap with simple sentences indicates that length alone cannot fully distinguish these two categories, an important consideration for classification systems relying heavily on surface features. Moreover, the compound sentence distribution shifts further right and broadens, typically spanning 10-18 words. This wider spread reflects the coordination of independent clauses, introducing variability in construction. Notably, its overlap with complex sentences suggests structural ambiguity in real-world data, where coordination and subordination may produce comparable sentence lengths despite differing grammatical relations. Furthermore, the compound-complex sentences demonstrate the most distinctive pattern, with a noticeably flatter and right-skewed distribution extending beyond 30 words. This indicates both higher average length and greater dispersion, consistent with their syntactic richness involving multiple clauses and clause types. The long tail highlights occasional highly elaborate constructions, which may pose challenges for both parsing and classification tasks.

In short, while sentence length correlates with structural complexity, it is not a definitive discriminator. The substantial overlaps among adjacent categories (simple-complex and complex-compound) suggest that structural classification cannot rely solely on length-based features. Instead, deeper syntactic or clause-level representations are necessary for robust differentiation. From a computational perspective, this distributional insight supports the need for models that capture hierarchical linguistic structure rather than depending on shallow metrics such as token count.



Similarly, Fig. \ref{fig:tkde} illustrates the KDE of sentence length distributions across three temporal categories. Unlike the structural distributions observed earlier, the temporal categories display a substantially higher degree of overlap, suggesting weaker differentiation based on sentence length. All three categories show a strong concentration within shorter length ranges, generally between 5-12 words. That means temporal reference does not inherently demand substantial variation in sentence length. However, subtle distributional distinctions are still observable. Future tense sentences demonstrate a slightly sharper peak within shorter lengths, typically around 6-9 words, suggesting a tendency toward more concise constructions. This may reflect the frequent use of auxiliary-based future marking, which allows temporal reference without significant syntactic expansion. In contrast, present tense sentences exhibit a broader and more right-skewed distribution, extending further into longer sentence ranges beyond 20 words. This suggests greater variability and flexibility in present-tense usage, possibly due to its dominance in descriptive, habitual, and ongoing contexts that accommodate elaboration. Past tense sentences occupy an intermediate position, showing moderate spread and overlap with both present and future categories. While their distribution resembles that of future tense in central tendency, the slightly longer tail indicates occasional narrative or descriptive expansions.
\\ \quad A key observation is the extensive overlap among all three temporal categories, implying that sentence length alone is a weak predictor of temporal classification. Unlike structural complexity, which tends to correlate with length, temporal reference is primarily expressed through morphological or auxiliary markers rather than clause expansion. From a computational standpoint, this suggests that surface-level metrics such as token count offer limited discriminative value for temporal classification tasks. Instead, models must rely on verb morphology, tense markers, and contextual cues to effectively distinguish temporal categories. 


\vspace{-1em}
\subsection{Baseline Evaluation}
To establish a foundational performance benchmark for both structural and temporal classification, we evaluate two broad modeling paradigms: multilingual encoders \cite{litschko2022cross} and dual-encoder architectures \cite{gupta2023dual}, where SMOTE \cite{khandaker2025handling} is applied to the embedding representations to mitigate class imbalance. The multilingual models (mBERT and XLM-R) serve as general-purpose baselines capable of handling cross-lingual representations, while the dual-encoder configurations combine English-centric contextual modeling (BERT/DistilBERT) with Bangla-specific pretrained encoders (BanglaBERTBase/BanglishBERT). 

The results in Table \ref{tab:bilingual_results} reveal several important patterns. First, multilingual encoders demonstrate reasonably strong baseline performance across both tasks, confirming their general effectiveness in low-resource and mixed-language scenarios. Among them, XLM-R consistently outperforms mBERT, achieving higher accuracy and F1-Scores for both structure and tense classification. This aligns with prior expectations, as XLM-R benefits from larger-scale multilingual pretraining and improved contextual representations. However, dual-encoder models exhibit a consistent performance advantage over multilingual baselines. The BERT + BanglishBERT configuration achieves the highest scores across all evaluation metrics, suggesting that combining complementary representations enhances the model’s sensitivity to linguistic nuances. Unlike multilingual encoders, which distribute representational capacity across many languages, dual encoders can more effectively capture language-specific syntactic and morphological patterns. The performance gap between full BERT-based dual encoders and DistilBERT-based variants further highlights the role of representational richness.

The performance gain is primarily attributed to language-specific representational specialization. While multilingual models share subword vocabularies across languages, this often leads to suboptimal tokenization for morphologically rich languages like Bangla. In contrast, dual encoders leverage dedicated tokenization schemes and pretrained representations, enabling better capture of syntactic and morphological nuances.

\vspace{-1em}
\section{Conclusion}
\label{sec:conclusion}
This paper presented BiST, a rigorously constructed Bangla-English bilingual corpus for sentence-level grammatical classification across structural and temporal dimensions. Designed to address the scarcity of reliable bilingual resources for low-resource settings, the corpus integrates systematic data collection, and multi-stage annotation with inter-annotator agreement to ensure consistency and reproducibility. Corpus-level analyses confirm realistic linguistic distributions and structural diversity, while baseline experiments demonstrate that dual-encoder architectures consistently outperform general multilingual encoders, highlighting the importance of language-specific representations. BiST thus establishes a unified and experimentally validated benchmark for bilingual grammatical modeling and cross-lingual research.


\vspace{-1em}
\bibliographystyle{lrec2026-natbib}
\bibliography{lrec2026-example}


\appendix

\section{Limitations and Future works}
Despite its contributions, the proposed BiST corpus has several limitations. The annotation scheme is coarse-grained, focusing only on structure and tense, and does not capture finer linguistic phenomena such as aspect, modality, or discourse relations. The corpus excludes code-switched instances, which may limit real-world variability. Additionally, while the corpus is moderately large, further expansion in size and domain diversity would improve its generalizability. 
In future work, we plan to expand the corpus with additional grammatical dimensions and linguistic phenomena, as well as explore code-mixed Bangla-English scenarios to better reflect real-world language use. We also aim to evaluate emerging multilingual foundation models on the benchmark to further advance low-resource grammatical understanding.
\vspace{-1em}
\section{Acknowledgment}
The authors acknowledge the use of ChatGPT (OpenAI) as a assistive tool for improving language, assisting with code, and organizing concepts. All generated material was rigorously reviewed, verified, and refined by the authors, who bear full responsibility for the accuracy, validity, and integrity of the manuscript.

\end{document}